\pdfoutput=1

\documentclass[11pt]{article}
\usepackage{ulem}
\usepackage{arydshln}
\usepackage{amssymb}
\usepackage{makecell}  
\usepackage{mathtools}
\usepackage{multirow}
\usepackage{xcolor}
\usepackage{booktabs}
\usepackage[preprint]{acl}
\usepackage{amsmath}
\usepackage{times}
\usepackage{latexsym}

\usepackage[T1]{fontenc}

\usepackage[utf8]{inputenc}

\usepackage{microtype}

\usepackage{inconsolata}

\usepackage{graphicx}
\usepackage{algorithm}
\usepackage{algpseudocode}
\usepackage{tcolorbox}
\usepackage{enumerate}
\usepackage{enumitem}

\title{Efficient Long CoT Reasoning in Small Language Models}

\makeatletter
\def\@fnsymbol#1{\ensuremath{\ifcase#1\or \dagger\or *\or \ddagger\or
   \mathsection\or \mathparagraph\or \|\or **\or \dagger\dagger
   \or \ddagger\ddagger \else\@ctrerr\fi}}
\makeatother

\def\myand{\end{tabular}\hss\egroup \hfil\hfil\egroup
           \hbox to \linewidth\bgroup\large \hfil\hfil
             \hbox to 0pt\bgroup\hss \begin{tabular}[t]{c}\bf}

\author{
  Zhaoyang Wang$^{1}$~\thanks{~Equal contribution.}~
  Jinqi Jiang$^{2}$~\footnotemark[1]~
  Tian Qiu$^{3}$~\footnotemark[1]~
  Hui Liu$^{4}$\myand
  Xianfeng Tang$^{4}$~ Huaxiu Yao$^{1}$\\
  $^{1}$University of North Carolina at Chapel Hill\\ 
  $^{2}$Huazhong University of Science and Technology~~ \\$^{3}$Fudan University~~ $^{4}$Amazon\\
  \texttt{\{zhaoyang,huaxiu\}@cs.unc.edu}
}

\begin{document}
\maketitle

\begin{abstract}
Recent large reasoning models such as DeepSeek-R1 exhibit strong complex problems solving abilities by generating long chain-of-thought (CoT) reasoning steps.
It is challenging to directly train small language models (SLMs) to emerge long CoT. Thus, distillation becomes a practical method to enable SLMs for such reasoning ability.
However, the long CoT often contains a lot of redundant contents (e.g., overthinking steps) which may make SLMs hard to learn considering their relatively poor capacity and generalization.
To address this issue, we propose a simple-yet-effective method to prune unnecessary steps in long CoT, and then employ an on-policy method for the SLM itself to curate valid and useful long CoT training data.
In this way, SLMs can effectively learn efficient long CoT reasoning and preserve competitive performance at the same time. 
Experimental results across a series of mathematical reasoning benchmarks demonstrate the effectiveness of the proposed method in distilling long CoT reasoning ability into SLMs which maintains the competitive performance but significantly reduces generating redundant reasoning steps.

\end{abstract}

\section{Introduction}

Chain-of-thought (CoT) prompting \cite{jason-wei-cot,wang2023selfconsistency,kojima2022large} has emerged as an effective approach for enhancing the reasoning capabilities of large language models (LLMs). Explicitly prompting models with phrases such as ``Let's think step by step'' and incorporating CoT-rich data during training have already become standard practices~\cite{meta2024llama3,abdin2024phi,qwen2.5,liu2024deepseek,team2024gemma}. Recent advancements in complex reasoning as OpenAI’s o1~\cite{OpenAI-O1}, Qwen-QwQ~\cite{qwq32b}, and DeepSeek-R1~\cite{deepseek-r1} large reasoning models further demonstrate that scaling up the length of CoT steps can significantly improve model performance in solving complex reasoning problems. 
While these efforts extend the boundaries of what LLMs can achieve, they also introduce new challenges to small language models (SLMs) with about 7B parameters which often use distillation methods to learn such long CoT reasoning~\cite{deepseek-r1,openr1}.

\begin{figure}[t]
  \centering
  \includegraphics[width=\columnwidth]{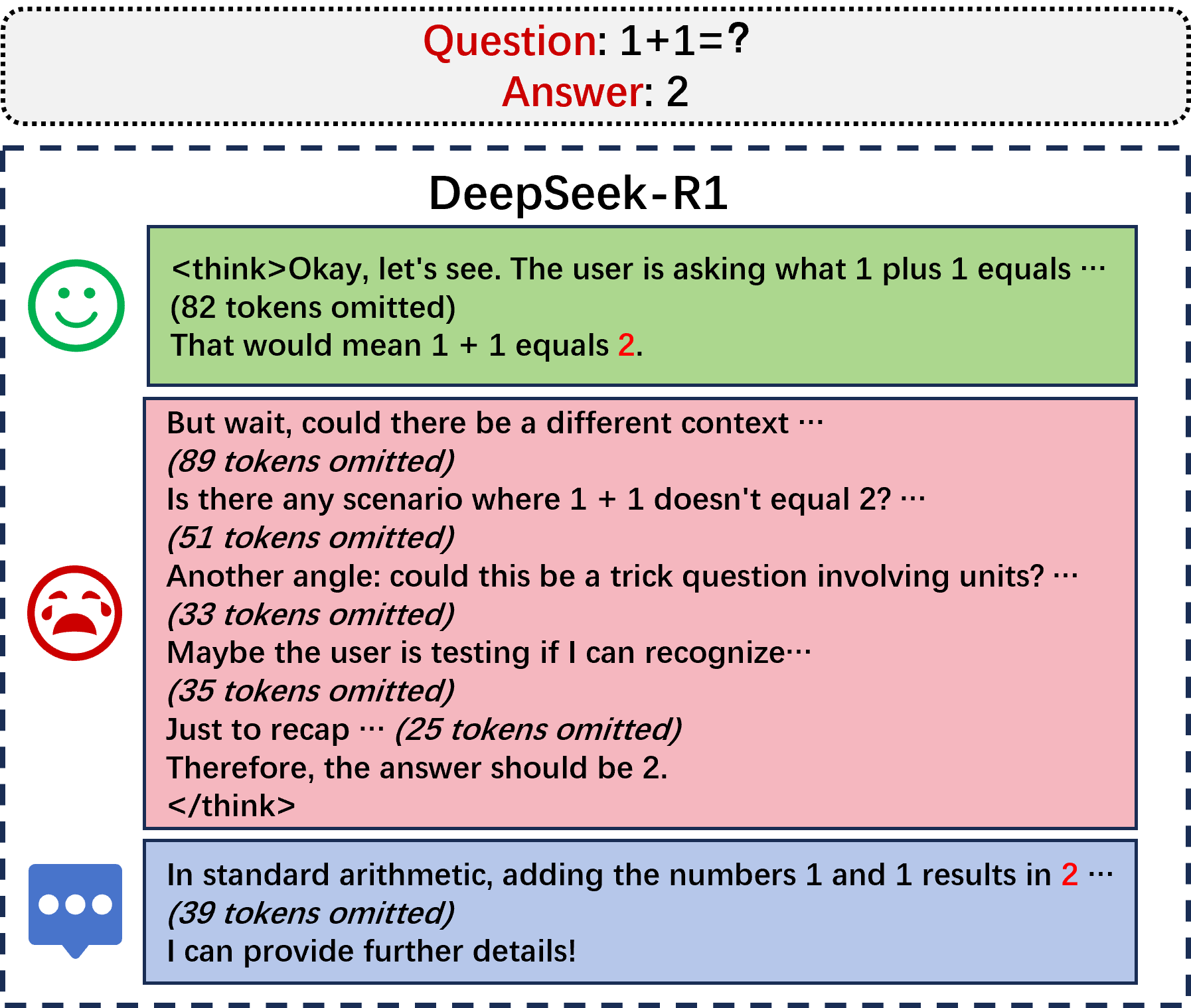} 
  \caption{
Illustration of redundant reasoning to a simple question by DeepSeek-R1~\citep{deepseek-r1}. The green part is sufficient for a correct answer, the red part is redundant, and the blue part is the summary response.}
  \label{fig:redundant-sample}
\end{figure}

While long CoT is necessary in scaling the performance of reasoning, its increasing length introduces significant computational inefficiency. 
Some recent works~\cite{tencent,l1-reasoning,yang2025think,LightThinker} have found that generated long CoT traces often contain many redundant reasoning steps even to the very simple question, as shown in Figure~\ref{fig:redundant-sample}. Those redundant reasoning steps may not only bring unnecessary computation burden during test time, but also affect the reasoning performance~\cite{sui2025stop,l1-reasoning,wu2025effectively,marjanovic2025deepseek}. Moreover, long CoT with redundant reasoning steps can make barriers to the distillation process, since SLMs have relatively poor capability and generalization.

To address this issue, existing works propose to use heuristic rules such as minimum reasoning length with correct final answer~\citep{tencent}, design length based rewards for reinforcement learning~\citep{l1-reasoning,yi2025shorterbetter,yang2025think}, or advanced prompting methods~\citep{wu2025effectively,munkhbat2025self,xia2025tokenskipcontrollablechainofthoughtcompression,han2025tokenbudgetawarellmreasoning,nayab2025concisethoughtsimpactoutput}. However, these methods either rely on re-designing the rewards during reinforcement learning which is often less effective than direct distillation from larger reasoning models~\citep{deepseek-r1,openr1} and requires more computation resources, or do not consider the target SLM's reasoning ability when selecting the long CoT training data. This leads to the question: \textit{How can high-quality CoT traces generated by large reasoning models be efficiently distilled into SLMs}?

In this paper, we first observed that long CoT reasoning generated by large reasoning models such as DeepSeek-R1~\citep{deepseek-r1} often contains unnecessary reasoning steps, even for the simple questions such as ``What is 1+1?''. This observation suggests that SLMs may not need to learn the entire CoT reasoning process, but rather focus on the essential reasoning steps that lead to the correct answer. 
Motivated by this, we propose a simple yet effective method to prune redundant reasoning steps in generated long CoT data. Specifically, We use binary cutting to efficiently search for the shortest concise CoT steps that lead to the correct answer, which greatly reduces the searching space and time complexity.
Further, we noticed that SLMs can often directly infer the correct answer from partial long CoT steps, and those segments of CoT vary with SLMs' own capabilities. Based on this, we propose an on-policy distillation method, in order to enhance the aforementioned binary cutting method to search the most tailored segments of long CoT for the target SLM.
Finally, we utilize these tailored concise CoT data to fine-tune the target SLM for complex reasoning with supervised fine-tuning (SFT) and direct preference optimization (DPO)~\citep{dpo}.
Experimental results across a series of mathematical reasoning tasks demonstrate the effectiveness of the proposed method in distilling long CoT reasoning ability into SLMs while significantly reducing the generation of redundant reasoning steps, which leads to a more efficient long CoT reasoning paradigm for SLMs.

In summary, our contributions are three-fold:

\begin{enumerate}[itemsep=2pt, parsep=0pt, partopsep=0pt, topsep=2pt, label=\arabic*)]
    \item We observed that long CoT reasoning generated by large reasoning models often contains unnecessary reasoning steps, which is then identified to be harmful for distilling such complex reasoning ability into SLMs.
    \item We propose a simple yet effective method to prune redundant reasoning steps in long CoT, which uses binary cutting for efficiently searching and on-policy validation that adapts to the target SLM's capability. 
    \item Experiments and analysis demonstrate the effectiveness of our method in enabling SLMs with efficient long CoT reasoning, which significantly reduces generating redundant reasoning steps while preserving performance. 
\end{enumerate}


\section{Related Work}

\paragraph{Chain of Thought.}
Chain of thought (CoT) reasoning has been widely adopted to enable LLMs to perform reasoning in a step-by-step manner~\citep{wei2022emergent,jason-wei-cot,kojima2022large,wang2023selfconsistency,zhou2023leasttomost}. Recently, the emergence of large reasoning models such as OpenAI's o1 series~\citep{OpenAI-O1}, DeepSeek-R1~\citep{deepseek-r1}, Qwen's QwQ~\citep{qwq32b} and Kimi's k1.5~\citep{team2025kimi} has demonstrated that scaling up the length of CoT reasoning can further improve model performance on complex reasoning tasks, which calls for the need to empower SLMs with such long CoT reasoning ability.

\paragraph{Redundancy in Long CoT.}
Large reasoning models often exhibit an overthinking problem, generating unnecessary or repetitive CoT steps that inflate sequence length which can lead to inefficiency and even harm final answer accuracy~\cite{sui2025stop,wu2025effectively,marjanovic2025deepseek,tencent}.  
To mitigate this issue, existing works have explored several approaches such as heuristic pruning methods that truncate CoT to the minimal prefix yielding a correct answer~\cite{tencent}, reinforcement-learning approaches that incorporate length-based penalties into the reward function~\cite{l1-reasoning,yang2025think,yi2025shorterbetter}, and alternative prompting techniques that guide models toward more concise reasoning~\cite{wu2025effectively,xia2025tokenskipcontrollablechainofthoughtcompression,han2025tokenbudgetawarellmreasoning,nayab2025concisethoughtsimpactoutput}. 
However, these methods often ignore the fact that different SLMs vary in their reasoning capabilities. In contrast, our approach combines a simple binary-cutting algorithm with an on-policy search tailored to the target SLM. This enables an efficient \(O(\log n)\) search for concise long CoT segments while adapting to the capability of the small model, ensuring both efficiency and performance in long CoT reasoning.

\paragraph{LLM Distillation.}
Distilling knowledge from LLMs into open-source SLMs has proven to be a simple yet effective approach to empower SLMs with new capabilities~\citep{hinton2015distillingknowledgeneuralnetwork,xu2024surveyknowledgedistillationlarge}. 
Beyond this, research community has successfully extended distillation to reasoning tasks by fine-tuning SLMs on rich CoT data annotated by LLMs~\citep{ho2022large,fuyaospecializing,magister2023teaching,wang2023democratizing,hsieh2023distilling}. 
More recently, OpenR1~\citep{openr1} and DeepSeek-R1's SLMs series~\citep{deepseek-r1} have demonstrated that SLMs can be effectively trained to perform complex reasoning via long CoT distillation. However, these efforts overlook the detrimental effects of redundant or unnecessary reasoning steps within the CoT process, which can overwhelm SLMs' limited capacity and generalization. In this paper, we focus on pruning such redundancy to enable efficient long CoT distillation and tailor the data to the target SLM.

\begin{figure*}[ht]
  \centering
  \includegraphics[width=1.0\textwidth]{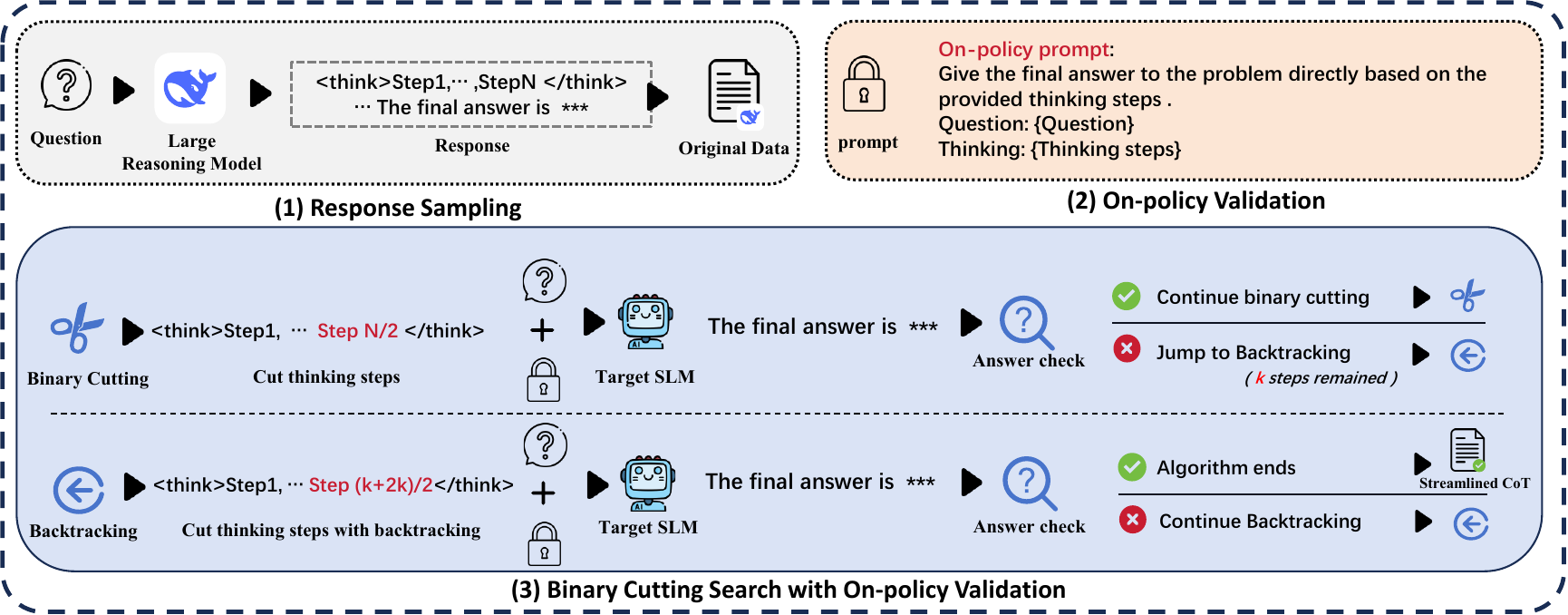}
  \caption {Overview of the proposed streamlining long CoT method which includes 3 key stages for data curation. (1) Response sampling that samples original long CoT reasoning samples from the large reasoning model. (2) On-policy Validation which prompts the target SLM to generate the final answer based on the segments of the reasoning (thinking). (3) Binary Cutting Search with On-policy Validation that combines binary cutting and on-policy validation to search valid streamlined long CoT reasoning steps, in order to fine-tune the target SLM.}
  \label{fig:main-figure}
\end{figure*}

\section{Method}
\subsection{Background}
Large reasoning models (e.g., DeepSeek-R1) are capable of generating long CoT reasoning steps to solve complex problems. However, these models usually produce excessive reasoning steps, where many of them are devoted to repeatedly verifying or reconfirming the correctness of an already correct answer. This behavior leads to unnecessarily long outputs and may also be harmful for the reasoning performance~\citep{marjanovic2025deepseek}.

The goal of this paper is to enable open-source SLMs to take advantage of long CoT reasoning ability while maintaining efficiency. To achieve this, most of existing works introduce length based penalties into the reward function~\cite{l1-reasoning,yang2025think,yi2025shorterbetter}, then train SLMs with similar reinforcement learning adopted by DeepSeek-R1~\citep{deepseek-r1}. However, this approach is often less effective than direct distillation from larger reasoning models~\citep{deepseek-r1} in terms of both performance and training efficiency. Thus, we choose to search segments of the long CoT reasoning that meet two constraints: (1) \textit{valid}, it should be sufficient for the target SLM to generate the correct final answer, and (ii) \textit{efficient}, it should be as short as possible to maximize efficiency. Then, we can use those long CoT segments to fine-tune the target SLM to achieve efficient long CoT reasoning.

In the following sections, we will first present the curation of streamlined long CoT data which includes the proposed binary cutting method and on-policy validation method, and then detail the fine-tuning process for the target SLM. The whole framework is illustrated in Figure~\ref{fig:main-figure}.

\subsection{Streamlining Long CoT}

\subsubsection{Response Sampling}
To collect long CoT data, we first sample a set of long CoT reasoning responses generated by the large reasoning models. Given the input question $Q$, the complete response $R$ includes two parts as shown in Figure~\ref{fig:redundant-sample}: (1) the thinking part $T$, which is the intermediate reasoning process enclosed within special \texttt{<think>} and \texttt{</think>} tags, and (2) the final response part, which is the summary paragraph to the thinking part and contains the final answer to the question.
The thinking part is often natural paragraph structure, and each paragraph can be viewed as a reasoning step. We split the thinking part $T$ into a list of reasoning steps $T = [s_1, s_2, \ldots, s_n]$, and we will operate on the step level for the subsequent methods. Combined with the ground truth final answer $A$, we can obtain a dataset of triplets $D_{\text{original}} = \{(Q_i, T_i, A_i)\}_{i=1}^N$. The objective is to find a subset of reasoning steps $T_{i}^{j:k}$ for each example $i$ such that maximizes the efficiency of the reasoning process while ensuring the correctness of the final answer, where $[j:k]$ means the set of steps $[s_j, s_{j+1}, \ldots, s_k]$.

\subsubsection{Binary Cutting}
Previous work~\citep{tencent} proposes First‐Correct Solutions (FCS), which truncates the reasoning process to the minimal prefix yielding a correct answer.  FCS requires checking every prefix $T^{1:k} = [s_1,\ldots,s_k]$ for $k=1,\dots,n$.  
While this linear search requires $O(n)$ complexity, an unconstrained search over all \(2^n\) subsequences would be intractable and break the reasoning chain’s coherence.  To maintain contiguity yet improve efficiency, we likewise restrict our search to prefixes, but replace the linear search with a binary cutting strategy.  This reduces the computation complexity to \(O(\log_{2} n)\) while still guaranteeing the discovery of the minimal valid prefix.
Specifically, at each iteration we compute $m = \bigl\lfloor (low + high)/2 \bigr\rfloor$, truncate the original reasoning chain \(T = [s_{1},\ldots,s_{n}]\) to the prefix \(T^{1:m}\), and invoke the validation function \(\phi(Q, T^{1:m}, A)\) which decides whether the segment is good and valid. If the prefix still yields the correct answer (\(\phi=1\)), we update \(best \leftarrow T^{1:m}\) and set \(high \leftarrow m\) to search even shorter prefixes. However, if the prefix fails to yield the correct answer (\(\phi=0\)), we introduce a backtracking mechanism to recover the last valid prefix: resetting \(low\) to the last mid‐point and perform a binary search upward toward \(n\), computing $m = \bigl\lceil (low + n)/2 \bigr\rceil$ at each step until \(\phi\) returns true again. The proposed binary cutting method with backtracking mechanism first aggressively cutting, which guarantees that the selected steps form a contiguous prefix, and recovers any essential steps that might have been over-pruned.
In contrast, as 61.83\% of the samples contain the answer in the last 10 steps, cutting from the beginning is not effective for SLM training.
This approach effectively reduces the search space and time complexity to $O(\log_2 n)$, while ensuring that the selected reasoning steps are both valid and efficient.

\begin{algorithm}[t]
\caption{Streamlining Long CoT.}
\label{alg:binary-reduction}
\begin{algorithmic}[1]
\Require Triplet $(Q, T, A)$ where $T=[s_{1},\dots,s_{n}]$, target model $M$, Validate function $\phi$
\Ensure Shortest valid contiguous long CoT segment $T^{1:k}$
\State $low \gets 1,\quad high \gets n,\quad best \gets T^{1:n}$
\While{$low < high$}  
    \State $mid \gets \lfloor (low + high) / 2 \rfloor$  
    \State $T' \gets T^{1:mid}$  
    \If{$\phi(Q, T', A, M)$ is \textbf{true}}  
        \State $best \gets T'$  
        \State $high \gets mid$    \Comment{can still shorten}  
    \Else  
        \State \textbf{break}      \Comment{enter backtracking}  
    \EndIf  
\EndWhile

\State $low \gets mid,\quad high \gets n$ \Comment{Backtracking: recover any over-pruned steps}
\While{$low < high$}
    \State $mid \gets \lceil (low + high) / 2 \rceil$
    \State $T' \gets T^{1:mid}$
    \If{$\phi(Q, T', A, M)$ is \textbf{true}}
        \State \Return $T'$      \Comment{found minimal valid prefix}
    \Else
        \State $low \gets mid + 1$
    \EndIf
\EndWhile

\State \Return $best$
\end{algorithmic}
\end{algorithm}

\subsubsection{On-Policy Validation}
Existing pruning methods such as FCS assume a single ``oracle'' validation criterion, typically provided by an additional judge model, to judge whether a truncated CoT remains correct. However, this ignores the fact that different SLMs exhibit distinct reasoning biases and strengths. To generate training data that is tailored to the target SLM’s own inductive preferences, we let the SLM $M$ itself serve as the validator in an on-policy paradigm.
Specifically, we construct a specialized prompt \(P_{\mathrm{policy}}\) (see Figure~\ref{fig:main-figure}, part (2)) which asks model $M$ to produce the final answer given only the question \(Q\) and a candidate prefix \(T^{1:k}\). We then can define the validation function $\phi$:
\begin{equation}
\scalebox{0.80}{$
\phi(Q,\,T^{1:k},\,A;\,M) = \mathbf{1}\!\bigl\{M_{t}(Q,\,T^{1:k},\,P_{\mathrm{policy}})=A\bigr\}
$}.
\end{equation}
During binary cutting, each prefix is accepted only if $\phi = 1$ under $M$. By relying on the target model’s own outputs rather than an external judge model, we ensure that the distilled CoT segments align with the SLM’s native reasoning capacity. This on-policy mechanism overlooked by prior methods yields a more coherent long CoT segment, since each retained prefix is one that the SLM can already interpret correctly. The whole method for streamlining long CoT  is illustrated in Algorithm~\ref{alg:binary-reduction}. See Appendix \ref{sec:appendix-prompts} for the full on-policy prompt.

\subsection{Fine-tuning SLM}
After applying binary cutting and on-policy validation methods, we obtain a distilled dataset $D_\text{distill} = \{(Q_i, R_i, Y_i)\}_{i=1}^{N}$, where $R$ is the original whole response, and $Y$ is the pruned concise response containing  the pruned thinking part and the final response part.
We can use this dataset $D_\text{distill}$ to fine-tune SLM to learn efficient long CoT reasoning via supervised fine-tuning (SFT) and direct preference optimization (DPO)~\citep{dpo}.

\subsubsection{SFT Training}
The most straightforward approach to leverage the obtained distilled data is to apply SFT training on the target model. Given the distilled dataset $D_\text{distill}$, the target SLM $M$ is fine-tuned to maximize the likelihood of the pruned reasoning $R$ conditioned on the input question $Q$ as follows:
\begin{equation}
    \mathcal{L}_\text{SFT} = - \mathbb{E}_{(Q, Y) \sim D_\text{distill}} \log M(Y|Q) .
    \label{eq:sft}
\end{equation}
After training, the SLM is expected to generate more concise and efficient reasoning steps while maintaining the correctness of the final answer.

\subsubsection{DPO Training}
To help the target model better distinguish between ``good'' and ``bad'' reasoning steps, we can leverage preference learning methods such as DPO to further fine-tune the model. Here, ``good'' refers to the pruned response with concise reasoning steps, while ``bad'' refers to the original response with redundant reasoning steps. The DPO training objective can be formulated as follows:
\begin{equation}
\begin{array}{l}
\quad \,\, \mathcal{L}_{\text{DPO}} = -\mathbb{E}_{(Q,R,Y) \sim \mathcal{D}_\text{distill}} \\[0.4em]
\scalebox{0.95}{$
\left[ \log \sigma
\left(
\beta \log \frac{M(Y | Q)}{M_{\text{ref}}(Y | Q)}
- \beta \log \frac{M(R | Q)}{M_{\text{ref}}(R | Q)}
\right) \right] $}
,
\end{array}
\label{eq:dpo}
\end{equation}
where $\sigma(\ast)$ denotes the logistic function, $\beta=0.1$ is a hyperparameter of DPO, and $M_\text{ref}$ is the frozen reference model typically the SLM after SFT training. Thanks to the significant difference between two responses $R$ and $Y$ in terms of the response length, the DPO training can effectively help the target SLM to learn the preference of concise reasoning steps over redundant ones.

\section{Experiments}
\subsection{Experimental Setup}

\paragraph{Datasets.}
To evaluate not only the reasoning performance but also the efficiency of long CoT reasoning, we benchmark SLMs on three mathematical reasoning datasets of increasing difficulty: GSM8K~\citep{GSM8K}, MATH~\citep{MATH}, and AIME~\citep{maaAIME2024}.
GSM8k is a primary school level mathematical dataset requiring basic arithmetic and logic. MATH we used is a widely used subset of original dataset which contains 500 challenging high school competition-level math problems. AIME consists of extremely difficult math problems spanning from 1983 to 2025, aiming to test model's generalization ability.

\paragraph{Models.}
We conducted experiments mainly on two popular open-source SLMs,  Llama-3.1-8B-Instruct~\citep{meta2024llama3} and Qwen2.5-7B-Instruct~\citep{qwen2.5}, which do not originally own the long CoT reasoning ability.

\paragraph{Baselines Methods.} 
In experiments, we mainly compare our method with the following baseline methods: (1) ``Base'', using the original SLM, as the baseline performance. (2) ``Full'', as described  in~\citet{deepseek-r1}, we directly use the original long CoT which may contain redundant reasoning steps for training, in order to directly show the necessity to remove such redundant steps. (3). ``Short CoT'', using the normal CoT data without scaling length to train the SLM, in order to demonstrate the effectiveness of scaling CoT. (4). ``FCS'', First‐Correct Solutions strategy~\citep{tencent} which linearly search the first occurred segment of long CoT with the correct final answer, in order to demonstrate the superiority of our method in pruning unnecessary reasoning steps and enhancing long CoT of SLMs. Following original implementations, we use a LLM Qwen2.5-14B-Instruct~\citep{qwen2.5} to segment the thinking part instead of nature line break.

\paragraph{Implementation Details.}
We use an existing large scale prepared long CoT data $D_\text{original}$ OpenR1-Math-220k~\citep{openr1}~\footnote{\scriptsize \href{https://huggingface.co/datasets/open-r1/OpenR1-Math-220k}{https://huggingface.co/datasets/open-r1/OpenR1-Math-220k}}, which has 220k math problems from NuminaMath 1.5~\citep{numina_math_datasets}, each paired with two to four long CoT reasoning generated by DeepSeek-R1~\citep{deepseek-r1}.
To form the distilled dataset, we perform binary cutting search with on-policy validation on the ``train'' split of this dataset which has about 93.7k samples. Finally, we have valid pruned long CoT data $D_\text{distill}$ about 25k samples for training. 
In most cases, the target SLM is fine-tuned by SFT training in 3 epochs, and DPO training in 1 epoch, with a learning rate of $1e-6$.   
We also noticed that single DPO training can also decrease the likelihood of the ``good'' response, thus we add the SFT loss with a weight of $0.3$ into Eq.~\ref{eq:dpo} for stable performance.
We use some regex expressions and format rules to extract the final answer from the generated response and calculate exact match accuracy.

\begin{table*}[t]
\centering
\caption{Main Results. Note that only Base and Short CoT report the full token counts, while all other methods count only tokens within \texttt{<think>...</think>}. The subscripts of Acc and Token indicate the percentage of decline or rise of the model relative to the model with SFT$_\text{Full}$ method.}

\label{tab:maintable}
\resizebox{\textwidth}{!}{

\begin{tabular}{c l cc cc cc}
\toprule
\multirow{2}{*}{\textbf{Model}} & \multirow{2}{*}{\textbf{Method}} & \multicolumn{2}{c}{\bf GSM8K} & \multicolumn{2}{c}{\bf MATH} & \multicolumn{2}{c}{\bf AIME} \\
\cmidrule(lr){3-4}\cmidrule(lr){5-6}\cmidrule(lr){7-8}
& & \bf Acc(\%)  & \bf \#Token & \bf Acc(\%)  & \bf \#Token & \bf Acc(\%)  & \bf \#Token \\
\midrule
\multirow{8}{*}{\centering Llama} 
& Base & 76.80 & 232 & 37.80 & 2454 & 12.97 & 6204 \\
& SFT$_\text{Full}$ & 89.01 & 1051 & 54.80 & 3274 & 16.51 & 7262 \\
& SFT$_\text{Short CoT}$ & 60.05{\color{red!80!black}$_{\text{(32.54\%$\downarrow$)}}$} & 314{\color{green!50!black}$_{\text{(70.12\%$\downarrow$)}}$} & 23.40{\color{red!80!black}$_{\text{(57.30\%$\downarrow$)}}$} & 1522{\color{green!50!black}$_{\text{(53.51\%$\downarrow$)}}$} & 4.82{\color{red!80!black}$_{\text{(70.81\%$\downarrow$)}}$} & 2269{\color{green!50!black}$_{\text{(68.76\%$\downarrow$)}}$} \\
& SFT$_\text{FCS}$ & 85.52{\color{red!80!black}$_{\text{(3.92\%$\downarrow$)}}$} & 728{\color{green!50!black}$_{\text{(30.73\%$\downarrow$)}}$} & 47.20{\color{red!80!black}$_{\text{(13.87\%$\downarrow$)}}$} & 1769{\color{green!50!black}$_{\text{(45.97\%$\downarrow$)}}$} & 10.08{\color{red!80!black}$_{\text{(38.95\%$\downarrow$)}}$} & 2874{\color{green!50!black}$_{\text{(60.42\%$\downarrow$)}}$} \\
& SFT+DPO$_\text{FCS}$ & 87.57{\color{red!80!black}$_{\text{(1.62\%$\downarrow$)}}$} & 598{\color{green!50!black}$_{\text{(43.10\%$\downarrow$)}}$} & 50.20{\color{red!80!black}$_{\text{(8.39\%$\downarrow$)}}$} & 1392{\color{green!50!black}$_{\text{(57.48\%$\downarrow$)}}$} & 13.40{\color{red!80!black}$_{\text{(18.84\%$\downarrow$)}}$} & 2135{\color{green!50!black}$_{\text{(70.60\%$\downarrow$)}}$} \\

& SFT$_\text{Ours}$ & 87.34{\color{red!80!black}$_{\text{(1.88\%$\downarrow$)}}$} & 502{\color{green!50!black}$_{\text{(52.24\%$\downarrow$)}}$} & 54.00{\color{red!80!black}$_{\text{(1.46\%$\downarrow$)}}$} & 2322{\color{green!50!black}$_{\text{(29.08\%$\downarrow$)}}$} & 18.01{\color{red!80!black}$_{\text{(9.09\%$\uparrow$)}}$} & 5480{\color{green!50!black}$_{\text{(24.54\%$\downarrow$)}}$} \\
& SFT+DPO$_\text{Ours}$ & 87.41{\color{red!80!black}$_{\text{(1.80\%$\downarrow$)}}$} & 339{\color{green!50!black}$_{\text{(67.75\%$\downarrow$)}}$} & 52.40{\color{red!80!black}$_{\text{(4.38\%$\downarrow$)}}$} & 1324{\color{green!50!black}$_{\text{(59.56\%$\downarrow$)}}$} & 17.90{\color{red!80!black}$_{\text{(8.42\%$\uparrow$)}}$} & 3779{\color{green!50!black}$_{\text{(47.96\%$\downarrow$)}}$} \\
\midrule
\multirow{8}{*}{\centering Qwen} 
& Base & 83.70 & 280 & 59.40 & 600 & 21.76 & 1121 \\
& SFT$_\text{Full}$ & 90.37 & 1011 & 64.00 & 2712 & 28.51 & 6330 \\
& SFT$_\text{Short CoT}$ & 64.67{\color{red!80!black}$_{\text{(28.44\%$\downarrow$)}}$} & 125{\color{green!50!black}$_{\text{(87.64\%$\downarrow$)}}$} & 43.20{\color{red!80!black}$_{\text{(32.50\%$\downarrow$)}}$} & 487{\color{green!50!black}$_{\text{(82.04\%$\downarrow$)}}$} & 13.61{\color{red!80!black}$_{\text{(52.26\%$\downarrow$)}}$} & 987{\color{green!50!black}$_{\text{(84.41\%$\downarrow$)}}$} \\
& SFT$_\text{FCS}$ & 56.79{\color{red!80!black}$_{\text{(37.16\%$\downarrow$)}}$} & 384{\color{green!50!black}$_{\text{(62.02\%$\downarrow$)}}$} & 30.40{\color{red!80!black}$_{\text{(52.50\%$\downarrow$)}}$} & 1032{\color{green!50!black}$_{\text{(61.95\%$\downarrow$)}}$} & 18.44{\color{red!80!black}$_{\text{(35.32\%$\downarrow$)}}$} & 2615{\color{green!50!black}$_{\text{(58.69\%$\downarrow$)}}$} \\
& SFT+DPO$_\text{FCS}$ & 78.24{\color{red!80!black}$_{\text{(13.42\%$\downarrow$)}}$} & 524{\color{green!50!black}$_{\text{(48.17\%$\downarrow$)}}$} & 49.40{\color{red!80!black}$_{\text{(22.81\%$\downarrow$)}}$} & 1914{\color{green!50!black}$_{\text{(29.42\%$\downarrow$)}}$} & 21.22{\color{red!80!black}$_{\text{(25.57\%$\downarrow$)}}$} & 4131{\color{green!50!black}$_{\text{(34.74\%$\downarrow$)}}$} \\

& SFT$_\text{Ours}$ & 89.16{\color{red!80!black}$_{\text{(1.34\%$\downarrow$)}}$} & 382{\color{green!50!black}$_{\text{(62.22\%$\downarrow$)}}$} & 61.20{\color{red!80!black}$_{\text{(4.38\%$\downarrow$)}}$} & 901{\color{green!50!black}$_{\text{(66.78\%$\downarrow$)}}$} & 25.51{\color{red!80!black}$_{\text{(10.52\%$\downarrow$)}}$} & 3021{\color{green!50!black}$_{\text{(52.27\%$\downarrow$)}}$} \\
& SFT+DPO$_\text{Ours}$ & 89.92{\color{red!80!black}$_{\text{(0.50\%$\downarrow$)}}$} & 278{\color{green!50!black}$_{\text{(72.50\%$\downarrow$)}}$} & 56.60{\color{red!80!black}$_{\text{(11.56\%$\downarrow$)}}$} & 489{\color{green!50!black}$_{\text{(81.97\%$\downarrow$)}}$} & 21.54{\color{red!80!black}$_{\text{(24.45\%$\downarrow$)}}$} & 1836{\color{green!50!black}$_{\text{(71.00\%$\downarrow$)}}$} \\
\bottomrule
\end{tabular}
}
\end{table*}
\begin{table}[t]
\centering
\caption{Average token usage and remaining ratio by different streamlining methods. Note that the discrepancy in original token counts arises because, to ensure comparable training data across methods, only data deemed valid by the on-policy method is reused by FCS. }
\label{tab:avg_token_ratio}
\resizebox{\columnwidth}{!}{ 
\begin{tabular}{l c c c c}
\toprule
\textbf{Method} & \textbf{Model} & \textbf{\#Orig} & \textbf{\#Remaining} & \textbf{Ratio} \\
\midrule
\multirow{2}{*}{FCS} 
& Llama   & 3659.45 & 2531.68 & 69.18\% \\
&  Qwen   & 3875.14 & 2695.19 & 69.55\% \\
\midrule
\multirow{2}{*}{Random} 
& Llama  & 4665.98 & 2333.51 & 50.01\% \\
& Qwen   & 4919.19 & 2466.74 & 50.15\% \\
\midrule
\multirow{2}{*}{Ours} 
& Llama  & 3660.35 & 2263.17 & 61.83\% \\
& Qwen   & 3875.96 & 1967.19 & 50.75\% \\

\bottomrule
\end{tabular}
}
\end{table}
\begin{figure}[t]
  \centering
  \includegraphics[width=\columnwidth]{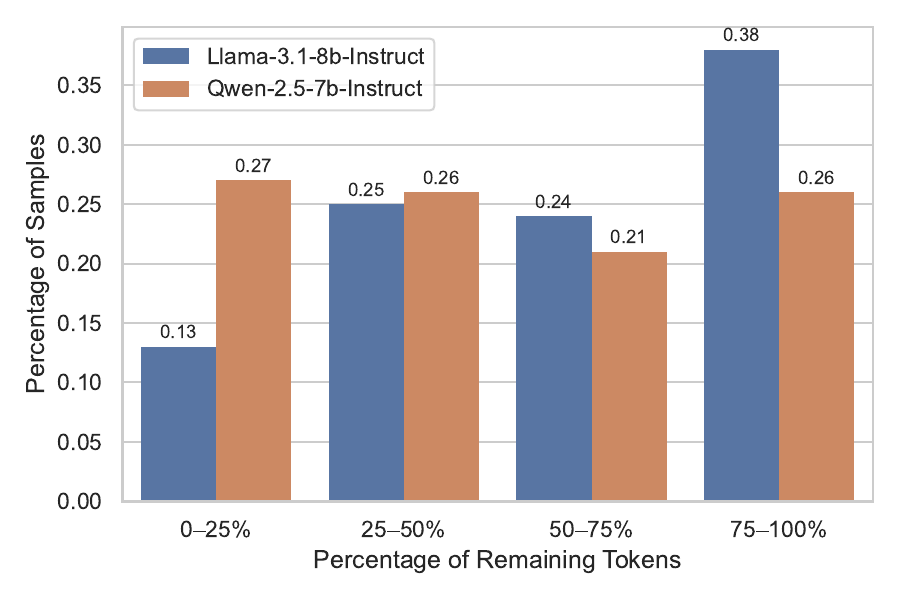} 
  \caption{Distribution of the remaining tokens ratio across different percentage intervals after streamlining.}
  \label{fig:percent-distribution}
\end{figure}

\subsection{Main Results}
Table~\ref{tab:maintable} shows the reasoning performance and efficiency of different methods.
\paragraph{Scaling CoT improves reasoning.}
All the long CoT based methods including ``Full'', ``FCS'' and ``Ours'' surpasses the model with ``short CoT'', which validates the effectiveness of scaling CoT length in mathematical reasoning.

\paragraph{Long CoT contains redundant steps.}
First, we shot that, for the easy reasoning task GSM8K, ``Full'' requires an average of 1051 tokens for reasoning, while does not show obvious better performance than other methods which often need less than half of the generation. 
Figure~\ref{fig:percent-distribution} clearly shows that about 40\% long CoT data can be streamlined over 50\% redundant reasoning steps by our method.
This supports the perspective that long CoT contains redundant steps especially for the easy tasks.
Then, compared to ``Full'', our method significantly decreases the number of generated thinking tokens, while only slightly decreases the reasoning performance for most cases.
For example, our Llama achieves 54.00\% accuracy on MATH after SFT, slightly lower than full long-CoT SFT (54.80\%), while reducing average number of tokens by 29.07\%.
Qwen model after distillation exhibits a modest decrease in accuracy compared to ``Full'' on AIME dataset, that might because Qwen is strong enough on this dataset as the base model.

\paragraph{Our method achieves efficient long CoT.}
Compared to the baseline method ``Full'' and another long CoT pruning method  ``FCS'', ours often show better reasoning performance than ``FCS'', while significantly increases the efficiency in long CoT reasoning. We also find that DPO training contributes the most to reducing the tokens, however, it harms the reasoning performance for hard tasks.
We further stat the results when at the same on-policy condition in Table~\ref{tab:avg_token_ratio}. For two target SLMs, Llama and Qwen, our method reduces the number of tokens to an average of 61.83\% and 50.75\%, respectively. However, the FCS and Random baseline methods retain more tokens, suggesting that the proposed binary cutting method is both efficient and effective in searching valid segments of long CoT reasoning. 
Additionally, We further evaluate the impacts of SFT loss weights in DPO training in Appendix~\ref{sec:DPO-various-settings}.

\section{Analysis}

\begin{table*}[t]
\centering
\caption{Ablation study results. ``Qwen data'' means that Llama is trained using streamlined long CoT data tailored for Qwen SLM. The subscripts of Acc and Token indicate the percentage of decline or rise of the model relative to SFT$_\text{Ours}$ and SFT+DPO$_\text{Ours}$ respectively.}
\label{tab:ablation-combined}
\resizebox{\textwidth}{!}{
\begin{tabular}{l l cc cc cc}
\toprule
\multirow{2}{*}{\textbf{Model}} & \multirow{2}{*}{\textbf{Method}} & \multicolumn{2}{c}{\bf GSM8K} & \multicolumn{2}{c}{\bf MATH} & \multicolumn{2}{c}{\bf AIME} \\
\cmidrule(lr){3-4}\cmidrule(lr){5-6}\cmidrule(lr){7-8}
& & \bf Acc (\%)  & \bf \#Token & \bf Acc (\%)  & \bf \#Token & \bf Acc (\%)  & \bf \#Token \\
\midrule
\multirow{6}{*}{Llama} 
& SFT$_\text{Ours}$ & 87.34 & 502 & 54.00 & 2322 & 18.01 & 5480 \\

& SFT$_\text{Random}$ & 
85.44{\color{red!80!black}$_{\text{(2.18\%$\downarrow$)}}$} & 
995{\color{green!50!black}$_{\text{(98.21\%$\uparrow$)}}$} & 
47.20{\color{red!80!black}$_{\text{(12.59\%$\downarrow$)}}$} & 
2894{\color{green!50!black}$_{\text{(24.63\%$\uparrow$)}}$} & 
8.68{\color{red!80!black}$_{\text{(51.81\%$\downarrow$)}}$} & 
5771{\color{green!50!black}$_{\text{(5.31\%$\uparrow$)}}$} \\

& SFT$_\text{Qwen data}$ & 
87.19{\color{red!80!black}$_{\text{(0.17\%$\downarrow$)}}$} & 
411{\color{green!50!black}$_{\text{(18.13\%$\downarrow$)}}$} & 
52.8{\color{red!80!black}$_{\text{(2.22\%$\downarrow$)}}$} & 
1896{\color{green!50!black}$_{\text{(18.35\%$\downarrow$)}}$} & 
16.08{\color{red!80!black}$_{\text{(10.72\%$\downarrow$)}}$} & 
5349{\color{green!50!black}$_{\text{(2.39\%$\downarrow$)}}$} \\
\cmidrule(lr){2-8}
& SFT+DPO$_\text{Ours}$ & 87.41 & 339 & 52.40 & 1324 & 17.90 & 3779 \\

& SFT+DPO$_\text{Random}$ & 
76.42{\color{red!80!black}$_{\text{(12.57\%$\downarrow$)}}$} & 
121{\color{green!50!black}$_{\text{(64.30\%$\downarrow$)}}$} & 
34.60{\color{red!80!black}$_{\text{(33.97\%$\downarrow$)}}$} & 
193{\color{green!50!black}$_{\text{(85.42\%$\downarrow$)}}$} & 
5.79{\color{red!80!black}$_{\text{(67.65\%$\downarrow$)}}$} & 
323{\color{green!50!black}$_{\text{(90.44\%$\downarrow$)}}$} \\

& SFT+DPO$_\text{Qwen data}$ & 
86.28{\color{red!80!black}$_{\text{(1.29\%$\downarrow$)}}$} & 
272{\color{green!50!black}$_{\text{(19.76\%$\downarrow$)}}$} & 
47.6{\color{red!80!black}$_{\text{(9.16\%$\downarrow$)}}$} & 
953{\color{green!50!black}$_{\text{(28.02\%$\downarrow$)}}$} & 
14.9{\color{red!80!black}$_{\text{(16.76\%$\downarrow$)}}$} & 
3006{\color{green!50!black}$_{\text{(20.46\%$\downarrow$)}}$} \\
\bottomrule
\end{tabular}
}
\end{table*}

\subsection{Ablation Study}
We conduct two types of ablation studies to evaluate the effectiveness of design choices of our method, shown in Table~\ref{tab:ablation-combined}.
First, we want to show that our binary cutting with backtracking mechanism can effectively identify redundant reasoning steps instead of randomly deleting. Thus, we introduce ``Random'' variant which randomly delete intermediate steps before applying either SFT or DPO training. From the results, we can find that random deletion leads to a substantial drop in accuracy across all datasets, indicating that preserving informative reasoning steps is essential for reasoning.
Second, to assess the role of our on-policy validation, we train Llama using concise CoT generated by the Qwen model. The performance is clearly inferior to our on-policy method, highlighting the importance of policy alignment between SLM's own reasoning capacity and long CoT data.

\begin{table}[t]
\centering
\caption{LLM judgment and human preference ranking on different long CoT data. ``T'', ``M'', and ``B'' represent top one, middle two, and bottom one, respectively.}
\label{tab:llm_judge_scores}
\begin{tabular}{l c c c c c}
\toprule
\textbf{Method} & \textbf{LLM} & \multicolumn{3}{c}{\textbf{Human}} \\
\cmidrule(lr){3-5}
 & & \textbf{T\%} & \textbf{M\%} & \textbf{B\%} \\
\midrule
Full     & 4.89 & 61 & 34 & 5 \\
\midrule
FCS               & 4.20 & 13 & 57 & 30 \\
Random     & 4.24 &  2 & 54 & 44 \\
Ours              & 4.44 & 24 & 55 & 21 \\
\bottomrule
\end{tabular}
\end{table}

\begin{figure*}[ht]
  \centering
  \includegraphics[width=1.0\textwidth]{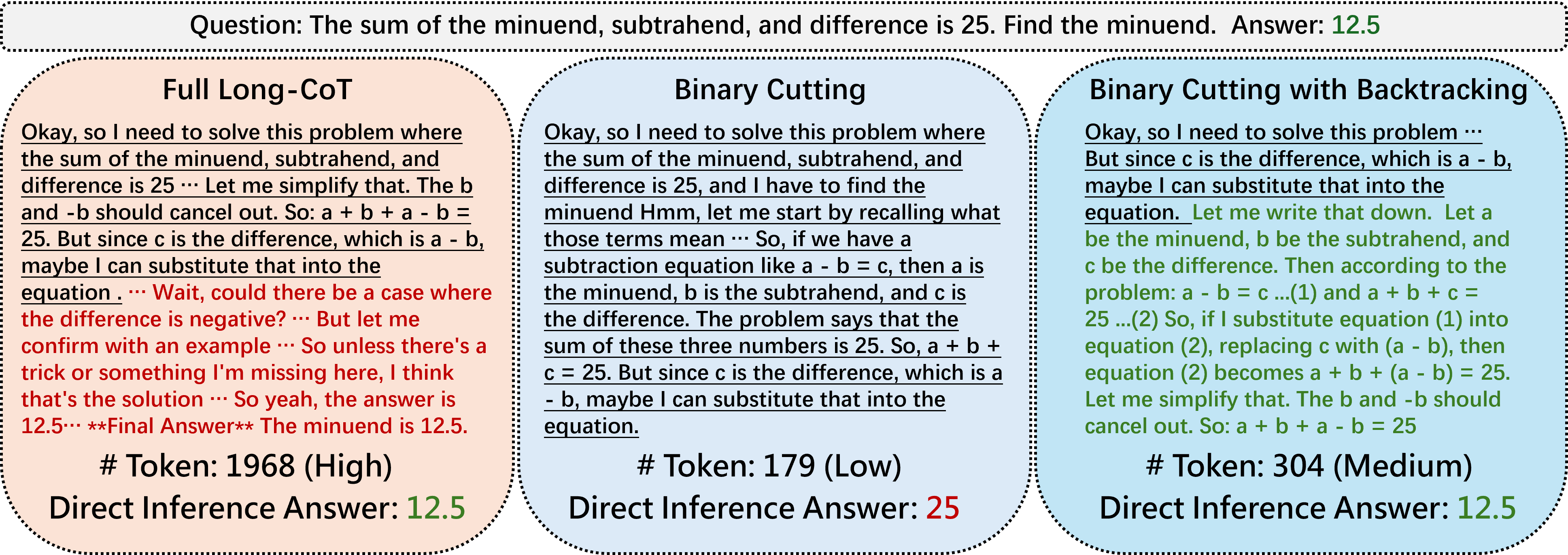}
  \caption {A case study of streamlining process of long CoT using our method. Underlined parts are the segments retained after binary cutting and are consistent across all three versions. Red text indicates steps removed during binary cutting. Green text marks reasoning steps restored in backtracking.}
  \label{fig:case-study-main}
\end{figure*}

\subsection{Quality of Streamlined CoT}

We use GPT-4.1 as a LLM-as-a-Judge to automatically evaluate the quality of CoT reasoning streamlined by different methods. A reward-based scoring prompt available at Appendix~\ref{sec:appendix-prompts} guides GPT-4.1 to rate each reasoning sample based on its correctness, completeness, conciseness and reasoning quality. We report the average scores from GPT-4.1 over 100 randomly selected examples. At the same time, we conduct human evaluation where annotators are asked to rank the four long CoT outputs into top-1 choice, middle two choice, and bottom one choice, based on overall reasoning quality and reasoning conciseness. This ranking scheme provides a coarse but interpretable assessment of human preference across different methods. 

As shown in Table~\ref{tab:llm_judge_scores}, we find that ``Full'' long CoT achieves the highest score due to its completeness. Among the other three streamlining methods, ``Random'' strategy performs poorly due to its natureal drawback in logical coherence, while FCS receives lower scores in conciseness. Our method, while less complete than ``Full'', achieves better balance between conciseness and logical consistency in reasoning.
Human evaluations are generally consistent with the LLM-based evaluation: ``Full'' method is most frequently ranked at the top one due to its completeness, while our method receives significantly more top and middle rankings than the other baseline methods, thanks to its backtracking mechanism and on-policy validation.

\subsection{Case Study}
Figure~\ref{fig:case-study-main} illustrates an example of streamlining a long CoT sample by our method with binary cutting and backtracking mechanism.
The original full long CoT can prompt the SLM to infer the correct answer directly, but it comes with the overthinking issue. After several rounds of binary cutting, the over-concise CoT is no longer sufficient to make SLM arrive at the correct final answer. We are surprising to find that in the backtracking stage, our method restores part of previously removed steps, which successfully guides the SLM to generate the correct answer. More examples are in Appendix~\ref{sec:examples-binary}.

\section{Conclusion}
In this paper, we tackle the challenge of distilling long CoT reasoning from large reasoning models into SLMs. We first identify that generated long CoT often introduces redundant or overthinking steps that waste computation during test time. To address this issue, we propose a binary cutting algorithm with backtracking, which locates the shortest contiguous prefix of the original reasoning that still yields a correct final answer in only $O(\log_2 n)$ complexity. Crucially, we further introduce on‐policy validation that uses the target SLM itself as the judge of whether a truncated segment of reasoning remains valid and useful for generating the final answer, thereby adapting the distilled data to the SLM’s own reasoning strengths. Extensive experiments on multiple math reasoning datasets demonstrate that our approach preserves competitive reasoning performance while significantly reducing the redundant tokens in long CoT reasoning. We believe these findings make meaningful contributions to efficient long CoT reasoning in SLMs.

\clearpage
\newpage
\section*{Limitations}
In this section, we discuss the limitations of our paper while offering potentially useful advice for future research.

\begin{enumerate}[itemsep=2pt, parsep=0pt, label=\arabic*)]
    \item The proposed method uses a binary-cutting algorithm for efficient pruning of long CoT traces. While this search strategy is not guaranteed to find the globally optimal subset of reasoning steps, we believe it offers significant practical advantages in terms of efficiency. Exploring more optimal yet efficient search algorithms remains an direction for future work.
    \item Our work focuses on the distillation scenario, which relies on a large reasoning model to provide high-quality reasoning traces. We do not consider reinforcement learning or self-training strategies for the SLM. While distillation offers high efficiency, alternative training paradigms are still valuable and complementary to our approach. 
    \item Due to limited computational resources, we evaluate our method only on two 7B-level models. We do not test across a wider range of model sizes. However, we believe 7B-scale models are the most commonly adopted among the community. Our framework should be applicable to any target model size.
\end{enumerate}

\bibliography{custom}

\begin{thebibliography}{39}
\providecommand{\natexlab}[1]{#1}

\bibitem[{Abdin et~al.(2024)Abdin, Aneja, Awadalla, Awadallah, Awan, Bach, Bahree, Bakhtiari, Bao, Behl et~al.}]{abdin2024phi}
Marah Abdin, Jyoti Aneja, Hany Awadalla, Ahmed Awadallah, Ammar~Ahmad Awan, Nguyen Bach, Amit Bahree, Arash Bakhtiari, Jianmin Bao, Harkirat Behl, et~al. 2024.
\newblock Phi-3 technical report: A highly capable language model locally on your phone.
\newblock \emph{arXiv preprint arXiv:2404.14219}.

\bibitem[{Aggarwal and Welleck(2025)}]{l1-reasoning}
Pranjal Aggarwal and Sean Welleck. 2025.
\newblock L1: Controlling how long a reasoning model thinks with reinforcement learning.
\newblock \emph{arXiv preprint arXiv:2503.04697}.

\bibitem[{Chen et~al.(2025)Chen, Xu, Liang, He, Pang, Yu, Song, Liu, Zhou, Zhang, Wang, Tu, Mi, and Yu}]{tencent}
Xingyu Chen, Jiahao Xu, Tian Liang, Zhiwei He, Jianhui Pang, Dian Yu, Linfeng Song, Qiuzhi Liu, Mengfei Zhou, Zhuosheng Zhang, Rui Wang, Zhaopeng Tu, Haitao Mi, and Dong Yu. 2025.
\newblock \href {https://arxiv.org/abs/2412.21187} {Do not think that much for 2+3=? on the overthinking of o1-like llms}.
\newblock \emph{Preprint}, arXiv:2412.21187.

\bibitem[{Cobbe et~al.(2021)Cobbe, Kosaraju, Bavarian, Chen, Jun, Kaiser, Plappert, Tworek, Hilton, Nakano, Hesse, and Schulman}]{GSM8K}
Karl Cobbe, Vineet Kosaraju, Mohammad Bavarian, Mark Chen, Heewoo Jun, Lukasz Kaiser, Matthias Plappert, Jerry Tworek, Jacob Hilton, Reiichiro Nakano, Christopher Hesse, and John Schulman. 2021.
\newblock \href {https://arxiv.org/abs/2110.14168} {Training verifiers to solve math word problems}.
\newblock \emph{Preprint}, arXiv:2110.14168.

\bibitem[{Face(2025)}]{openr1}
Hugging Face. 2025.
\newblock \href {https://github.com/huggingface/open-r1} {Open r1: A fully open reproduction of deepseek-r1}.

\bibitem[{Fu et~al.(2023)Fu, Peng, Ou, Sabharwal, and Khot}]{fuyaospecializing}
Yao Fu, Hao Peng, Litu Ou, Ashish Sabharwal, and Tushar Khot. 2023.
\newblock \href {https://proceedings.mlr.press/v202/fu23d.html} {Specializing smaller language models towards multi-step reasoning}.
\newblock In \emph{Proceedings of the 40th International Conference on Machine Learning}, volume 202 of \emph{Proceedings of Machine Learning Research}, pages 10421--10430. PMLR.

\bibitem[{Guo et~al.(2025)Guo, Yang, Zhang, Song, and Zhang}]{deepseek-r1}
Daya Guo, Dejian Yang, Haowei Zhang, Junxiao Song, and Ruoyu Zhang. 2025.
\newblock \href {https://arxiv.org/abs/2501.12948} {Deepseek-r1: Incentivizing reasoning capability in llms via reinforcement learning}.
\newblock \emph{Preprint}, arXiv:2501.12948.

\bibitem[{Han et~al.(2025)Han, Wang, Fang, Zhao, Ma, and Chen}]{han2025tokenbudgetawarellmreasoning}
Tingxu Han, Zhenting Wang, Chunrong Fang, Shiyu Zhao, Shiqing Ma, and Zhenyu Chen. 2025.
\newblock \href {https://arxiv.org/abs/2412.18547} {Token-budget-aware llm reasoning}.
\newblock \emph{Preprint}, arXiv:2412.18547.

\bibitem[{Hendrycks et~al.(2021)Hendrycks, Burns, Kadavath, Arora, Basart, Tang, Song, and Steinhardt}]{MATH}
Dan Hendrycks, Collin Burns, Saurav Kadavath, Akul Arora, Steven Basart, Eric Tang, Dawn Song, and Jacob Steinhardt. 2021.
\newblock \href {https://arxiv.org/abs/2103.03874} {Measuring mathematical problem solving with the math dataset}.
\newblock \emph{Preprint}, arXiv:2103.03874.

\bibitem[{Hinton et~al.(2015)Hinton, Vinyals, and Dean}]{hinton2015distillingknowledgeneuralnetwork}
Geoffrey Hinton, Oriol Vinyals, and Jeff Dean. 2015.
\newblock \href {https://arxiv.org/abs/1503.02531} {Distilling the knowledge in a neural network}.
\newblock \emph{Preprint}, arXiv:1503.02531.

\bibitem[{Ho et~al.(2023)Ho, Schmid, and Yun}]{ho2022large}
Namgyu Ho, Laura Schmid, and Se-Young Yun. 2023.
\newblock \href {https://doi.org/10.18653/v1/2023.acl-long.830} {Large language models are reasoning teachers}.
\newblock In \emph{Proceedings of the 61st Annual Meeting of the Association for Computational Linguistics (Volume 1: Long Papers)}, pages 14852--14882, Toronto, Canada. Association for Computational Linguistics.

\bibitem[{Kojima et~al.(2022)Kojima, Gu, Reid, Matsuo, and Iwasawa}]{kojima2022large}
Takeshi Kojima, Shixiang~Shane Gu, Machel Reid, Yutaka Matsuo, and Yusuke Iwasawa. 2022.
\newblock \href {https://openreview.net/forum?id=e2TBb5y0yFf} {Large language models are zero-shot reasoners}.
\newblock In \emph{Advances in Neural Information Processing Systems}.

\bibitem[{LI et~al.(2024)LI, Beeching, Tunstall, Lipkin, Soletskyi, Huang, Rasul, Yu, Jiang, Shen, Qin, Dong, Zhou, Fleureau, Lample, and Polu}]{numina_math_datasets}
Jia LI, Edward Beeching, Lewis Tunstall, Ben Lipkin, Roman Soletskyi, Shengyi~Costa Huang, Kashif Rasul, Longhui Yu, Albert Jiang, Ziju Shen, Zihan Qin, Bin Dong, Li~Zhou, Yann Fleureau, Guillaume Lample, and Stanislas Polu. 2024.
\newblock Numinamath.
\newblock \url{[https://huggingface.co/AI-MO/NuminaMath-1.5](https://github.com/project-numina/aimo-progress-prize/blob/main/report/numina_dataset.pdf)}.

\bibitem[{Liu et~al.(2024)Liu, Feng, Xue, Wang, Wu, Lu, Zhao, Deng, Zhang, Ruan et~al.}]{liu2024deepseek}
Aixin Liu, Bei Feng, Bing Xue, Bingxuan Wang, Bochao Wu, Chengda Lu, Chenggang Zhao, Chengqi Deng, Chenyu Zhang, Chong Ruan, et~al. 2024.
\newblock Deepseek-v3 technical report.
\newblock \emph{arXiv preprint arXiv:2412.19437}.

\bibitem[{{MAA}(2024)}]{maaAIME2024}
{MAA}. 2024.
\newblock \href {https://maa.org/math-competitions/american-invitational-mathematics-examination-aime} {American invitational mathematics examination --- aime}.
\newblock American Invitational Mathematics Examination -- AIME 2024, February 2024.

\bibitem[{Magister et~al.(2023)Magister, Mallinson, Adamek, Malmi, and Severyn}]{magister2023teaching}
Lucie~Charlotte Magister, Jonathan Mallinson, Jakub Adamek, Eric Malmi, and Aliaksei Severyn. 2023.
\newblock \href {https://doi.org/10.18653/v1/2023.acl-short.151} {Teaching small language models to reason}.
\newblock In \emph{Proceedings of the 61st Annual Meeting of the Association for Computational Linguistics (Volume 2: Short Papers)}, pages 1773--1781, Toronto, Canada. Association for Computational Linguistics.

\bibitem[{Marjanovi{\'c} et~al.(2025)Marjanovi{\'c}, Patel, Adlakha, Aghajohari, BehnamGhader, Bhatia, Khandelwal, Kraft, Krojer, L{\`u} et~al.}]{marjanovic2025deepseek}
Sara~Vera Marjanovi{\'c}, Arkil Patel, Vaibhav Adlakha, Milad Aghajohari, Parishad BehnamGhader, Mehar Bhatia, Aditi Khandelwal, Austin Kraft, Benno Krojer, Xing~Han L{\`u}, et~al. 2025.
\newblock Deepseek-r1 thoughtology: Let's< think> about llm reasoning.
\newblock \emph{arXiv preprint arXiv:2504.07128}.

\bibitem[{Meta-AI(2024)}]{meta2024llama3}
Meta-AI. 2024.
\newblock Introducing meta llama 3: The next generation of open models.
\newblock \url{https://ai.meta.com/blog/meta-llama-3-1/}.
\newblock Accessed: 2025-05-18.

\bibitem[{Munkhbat et~al.(2025)Munkhbat, Ho, Kim, Yang, Kim, and Yun}]{munkhbat2025self}
Tergel Munkhbat, Namgyu Ho, Seo~Hyun Kim, Yongjin Yang, Yujin Kim, and Se-Young Yun. 2025.
\newblock Self-training elicits concise reasoning in large language models.
\newblock \emph{arXiv preprint arXiv:2502.20122}.

\bibitem[{Nayab et~al.(2025)Nayab, Rossolini, Simoni, Saracino, Buttazzo, Manes, and Giacomelli}]{nayab2025concisethoughtsimpactoutput}
Sania Nayab, Giulio Rossolini, Marco Simoni, Andrea Saracino, Giorgio Buttazzo, Nicolamaria Manes, and Fabrizio Giacomelli. 2025.
\newblock \href {https://arxiv.org/abs/2407.19825} {Concise thoughts: Impact of output length on llm reasoning and cost}.
\newblock \emph{Preprint}, arXiv:2407.19825.

\bibitem[{OpenAI.(2024)}]{OpenAI-O1}
OpenAI. 2024.
\newblock \href {https://openai.com/index/ learning-to-reason-with-llms} {Learning to reason with llms.}
\newblock Accessed: 2025-03-05.

\bibitem[{Qwen-Team(2024)}]{qwen2.5}
Qwen-Team. 2024.
\newblock \href {https://qwenlm.github.io/blog/qwen2.5/} {Qwen2.5: A party of foundation models}.

\bibitem[{Rafailov et~al.(2023)Rafailov, Sharma, Mitchell, Manning, Ermon, and Finn}]{dpo}
Rafael Rafailov, Archit Sharma, Eric Mitchell, Christopher~D Manning, Stefano Ermon, and Chelsea Finn. 2023.
\newblock Direct preference optimization: Your language model is secretly a reward model.
\newblock \emph{Advances in Neural Information Processing Systems}, 36:53728--53741.

\bibitem[{Shridhar et~al.(2023)Shridhar, Stolfo, and Sachan}]{hsieh2023distilling}
Kumar Shridhar, Alessandro Stolfo, and Mrinmaya Sachan. 2023.
\newblock \href {https://doi.org/10.18653/v1/2023.findings-acl.441} {Distilling reasoning capabilities into smaller language models}.
\newblock In \emph{Findings of the Association for Computational Linguistics: ACL 2023}, pages 7059--7073, Toronto, Canada. Association for Computational Linguistics.

\bibitem[{Sui et~al.(2025)Sui, Chuang, Wang, Zhang, Zhang, Yuan, Liu, Wen, Zhong, Chen et~al.}]{sui2025stop}
Yang Sui, Yu-Neng Chuang, Guanchu Wang, Jiamu Zhang, Tianyi Zhang, Jiayi Yuan, Hongyi Liu, Andrew Wen, Shaochen Zhong, Hanjie Chen, et~al. 2025.
\newblock Stop overthinking: A survey on efficient reasoning for large language models.
\newblock \emph{arXiv preprint arXiv:2503.16419}.

\bibitem[{Team et~al.(2024)Team, Riviere, Pathak, Sessa, Hardin, Bhupatiraju, Hussenot, Mesnard, Shahriari, Ram{\'e} et~al.}]{team2024gemma}
Gemma Team, Morgane Riviere, Shreya Pathak, Pier~Giuseppe Sessa, Cassidy Hardin, Surya Bhupatiraju, L{\'e}onard Hussenot, Thomas Mesnard, Bobak Shahriari, Alexandre Ram{\'e}, et~al. 2024.
\newblock Gemma 2: Improving open language models at a practical size.
\newblock \emph{arXiv preprint arXiv:2408.00118}.

\bibitem[{Team(2025{\natexlab{a}})}]{team2025kimi}
Kimi Team. 2025{\natexlab{a}}.
\newblock \href {https://arxiv.org/abs/2501.12599} {Kimi k1.5: Scaling reinforcement learning with llms}.
\newblock \emph{Preprint}, arXiv:2501.12599.

\bibitem[{Team(2025{\natexlab{b}})}]{qwq32b}
Qwen Team. 2025{\natexlab{b}}.
\newblock \href {https://qwenlm.github.io/blog/qwq-32b/} {Qwq-32b: Embracing the power of reinforcement learning}.

\bibitem[{Wang et~al.(2023{\natexlab{a}})Wang, Wei, Schuurmans, Le, Chi, Narang, Chowdhery, and Zhou}]{wang2023selfconsistency}
Xuezhi Wang, Jason Wei, Dale Schuurmans, Quoc~V Le, Ed~H. Chi, Sharan Narang, Aakanksha Chowdhery, and Denny Zhou. 2023{\natexlab{a}}.
\newblock \href {https://openreview.net/forum?id=1PL1NIMMrw} {Self-consistency improves chain of thought reasoning in language models}.
\newblock In \emph{International Conference on Learning Representations}.

\bibitem[{Wang et~al.(2023{\natexlab{b}})Wang, Huang, Liu, Wang, Song, Zhang, Huang, Wei, Deng, Sun, and Zhang}]{wang2023democratizing}
Zhaoyang Wang, Shaohan Huang, Yuxuan Liu, Jiahai Wang, Minghui Song, Zihan Zhang, Haizhen Huang, Furu Wei, Weiwei Deng, Feng Sun, and Qi~Zhang. 2023{\natexlab{b}}.
\newblock \href {https://arxiv.org/abs/2310.13332} {Democratizing reasoning ability: Tailored learning from large language model}.
\newblock \emph{Preprint}, arXiv:2310.13332.

\bibitem[{Wei et~al.(2022{\natexlab{a}})Wei, Tay, Bommasani, Raffel, Zoph, Borgeaud, Yogatama, Bosma, Zhou, Metzler, Chi, Hashimoto, Vinyals, Liang, Dean, and Fedus}]{wei2022emergent}
Jason Wei, Yi~Tay, Rishi Bommasani, Colin Raffel, Barret Zoph, Sebastian Borgeaud, Dani Yogatama, Maarten Bosma, Denny Zhou, Donald Metzler, Ed~H. Chi, Tatsunori Hashimoto, Oriol Vinyals, Percy Liang, Jeff Dean, and William Fedus. 2022{\natexlab{a}}.
\newblock \href {https://openreview.net/forum?id=yzkSU5zdwD} {Emergent abilities of large language models}.
\newblock \emph{Transactions on Machine Learning Research}.
\newblock Survey Certification.

\bibitem[{Wei et~al.(2022{\natexlab{b}})Wei, Wang, Schuurmans, Bosma, Ichter, Xia, Chi, Le, and Zhou}]{jason-wei-cot}
Jason Wei, Xuezhi Wang, Dale Schuurmans, Maarten Bosma, Brian Ichter, Fei Xia, Ed~H. Chi, Quoc~V. Le, and Denny Zhou. 2022{\natexlab{b}}.
\newblock Chain-of-thought prompting elicits reasoning in large language models.
\newblock In \emph{Advances in Neural Information Processing Systems}.

\bibitem[{Wu et~al.(2025)Wu, Xiang, Wang, and Mittal}]{wu2025effectively}
Tong Wu, Chong Xiang, Jiachen~T Wang, and Prateek Mittal. 2025.
\newblock Effectively controlling reasoning models through thinking intervention.
\newblock \emph{arXiv preprint arXiv:2503.24370}.

\bibitem[{Xia et~al.(2025)Xia, Li, Leong, Wang, and Li}]{xia2025tokenskipcontrollablechainofthoughtcompression}
Heming Xia, Yongqi Li, Chak~Tou Leong, Wenjie Wang, and Wenjie Li. 2025.
\newblock \href {https://arxiv.org/abs/2502.12067} {Tokenskip: Controllable chain-of-thought compression in llms}.
\newblock \emph{Preprint}, arXiv:2502.12067.

\bibitem[{Xu et~al.(2024)Xu, Li, Tao, Shen, Cheng, Li, Xu, Tao, and Zhou}]{xu2024surveyknowledgedistillationlarge}
Xiaohan Xu, Ming Li, Chongyang Tao, Tao Shen, Reynold Cheng, Jinyang Li, Can Xu, Dacheng Tao, and Tianyi Zhou. 2024.
\newblock \href {https://arxiv.org/abs/2402.13116} {A survey on knowledge distillation of large language models}.
\newblock \emph{Preprint}, arXiv:2402.13116.

\bibitem[{Yang et~al.(2025)Yang, Lin, and Yu}]{yang2025think}
Junjie Yang, Ke~Lin, and Xing Yu. 2025.
\newblock Think when you need: Self-adaptive chain-of-thought learning.
\newblock \emph{arXiv preprint arXiv:2504.03234}.

\bibitem[{Yi and Wang(2025)}]{yi2025shorterbetter}
Jingyang Yi and Jiazheng Wang. 2025.
\newblock Shorterbetter: Guiding {Reasoning} {Models} to {Find} {Optimal} {Inference} {Length} for {Efficient} {Reasoning}.
\newblock \emph{arXiv}.

\bibitem[{Zhang et~al.(2025)Zhang, Zhu, Sun, Luo, Qiao, Du, Zheng, Chen, and Zhang}]{LightThinker}
Jintian Zhang, Yuqi Zhu, Mengshu Sun, Yujie Luo, Shuofei Qiao, Lun Du, Da~Zheng, Huajun Chen, and Ningyu Zhang. 2025.
\newblock \href {https://arxiv.org/abs/2502.15589} {Lightthinker: Thinking step-by-step compression}.
\newblock \emph{Preprint}, arXiv:2502.15589.

\bibitem[{Zhou et~al.(2023)Zhou, Sch{\"a}rli, Hou, Wei, Scales, Wang, Schuurmans, Cui, Bousquet, Le, and Chi}]{zhou2023leasttomost}
Denny Zhou, Nathanael Sch{\"a}rli, Le~Hou, Jason Wei, Nathan Scales, Xuezhi Wang, Dale Schuurmans, Claire Cui, Olivier Bousquet, Quoc~V Le, and Ed~H. Chi. 2023.
\newblock \href {https://openreview.net/forum?id=WZH7099tgfM} {Least-to-most prompting enables complex reasoning in large language models}.
\newblock In \emph{International Conference on Learning Representations}.

\end{thebibliography}

\appendix
\onecolumn
\section{Release Plan}\label{sec:release}
We will make our codebase, prompts, models and  datasets publicly available after the camera-ready deadline to facilitate reproducibility and further research.

\section{DPO Training Settings}\label{sec:DPO-various-settings}
Table~\ref{tab:dpo-various-settings} presents the hyperparameters used in DPO training under different settings, including loss weights and data configurations.
\begin{table*}[htbp]

\centering
\caption{Experimental results of DPO with different SFT loss weights and data sizes.}
\label{tab:dpo-various-settings}
\resizebox{\textwidth}{!}{
\begin{tabular}{c l cc cc cc}
\toprule
\multirow{2}{*}{\textbf{Model+Method}} & \multirow{2}{*}{\textbf{SFT\_weight / \#Data}} & \multicolumn{2}{c}{\bf GSM8K} & \multicolumn{2}{c}{\bf MATH} & \multicolumn{2}{c}{\bf AIME} \\
\cmidrule(lr){3-4}\cmidrule(lr){5-6}\cmidrule(lr){7-8}
& & \bf Acc (\%)  & \bf \#Token & \bf Acc (\%)  & \bf \#Token & \bf Acc (\%)  & \bf \#Token \\
\midrule
\multirow{7}{*}{\centering Llama+DPO} 
& 0.1 / 5K & 87.34 & 299 & 48.40 & 1102 & 15.65 & 2905 \\
& 0.2 / 5K & 85.22 & 345 & 46.20 & 1397 & 14.79 & 3444 \\
& 0.3 / 5K & 87.41 & 339 & 52.40 & 1324 & 17.90 & 3779 \\
& 0.1 / 10K & 88.02 & 387 & 50.00 & 1399 & 17.04 & 3972 \\
& 0.1 / 20K & 86.43 & 396 & 51.00 & 1651 & 17.90 & 4708 \\

\midrule
\multirow{5}{*}{\centering Qwen+DPO} 
& 0.1 / 5K & 88.32 & 171 & 54.80 & 247 & 18.22 & 627 \\
& 0.2 / 5K & 88.78 & 220 & 58.00 & 401 & 18.11 & 1078 \\
& 0.3 / 5K & 89.23 & 251 & 55.60 & 443 & 19.94 & 1353 \\
& 0.1 / 10K & 88.17 & 250 & 58.60 & 457 & 19.08 & 1409 \\
& 0.1 / 20K & 89.92 & 278 & 56.60 & 489 & 21.54 & 1836 \\
\bottomrule
\end{tabular}
}
\end{table*}

\section{Prompt Examples}\label{sec:appendix-prompts}
We provide representative prompt examples used in different stages of our pipeline, including on-policy generation and evaluation with LLM-as-a-Judge.

\paragraph{On-policy Prompt.}
Figure~\ref{fig:on-policy-answering-prompt} shows a typical prompt used to generate a final answer directly.
\begin{figure*}[htbp]
\centering
\begin{tcolorbox}[colback=gray!00, 
                  colframe=black, 
                  width=16cm, 
                  arc=1.5mm, auto outer arc,
                  left=0.9mm, right=0.9mm,
                  boxrule=0.9pt, colbacktitle = black!65!black
                 ]
                 
\textbf{On-policy answering prompt} \\
Given a mathematical problem and a series of thinking steps, your task is to determine the final answer based on the provided steps.  

Instructions:  

- Follow the given thinking steps exactly to derive the answer.

- Do not add any explanations, reasoning, or modifications.  

- Output only the final answer in the specified format.  

Formatting Requirement: 

- The output should be strictly formatted as:  

\#\#\#Answer: <final\_answer>

Now, here is the problem:  
\{question\}  

The thinking steps are given below:  
\{thinking\_steps\}  

\end{tcolorbox}
\caption{On-policy answering prompt format.}
\label{fig:on-policy-answering-prompt}
\end{figure*}

\paragraph{LLM-as-a-Judge Prompt.}
Figure~\ref{fig:llm-judge-prompt} shows how we guide the LLM to score CoT based on predefined rules.
\begin{figure*}[htbp]
\centering
\begin{tcolorbox}[colback=gray!00, 
                  colframe=black, 
                  width=16cm, 
                  arc=1.5mm, auto outer arc,
                  left=0.9mm, right=0.9mm,
                  boxrule=0.9pt, colbacktitle = black!65!black
                 ]
                 
\textbf{LLM-as-a-judge prompt} \\
Review the user's question and the corresponding chain-of-thought (CoT) reasoning using the additive 5-point scoring system described below. Points are accumulated based on the satisfaction of each criterion:

- Add 1 point if the CoT is relevant and initiates a reasoning process related to the user's inquiry, even if it is incomplete or contains some irrelevant steps.

- Add another point if the CoT addresses a substantial portion of the reasoning needed for the question, but does not completely resolve it or contains gaps in logic.

- Award a third point if the CoT outlines the basic logical structure needed to solve the problem, demonstrating generally sound reasoning without unnecessary tangents.

- Grant a fourth point if the CoT is clearly and logically structured, with each step building on the last. The reasoning should be easy to follow, focused on solving the user's question, and reasonably efficient, though not necessarily optimal in brevity or precision. It's acceptable to stop the reasoning early if the remaining steps are obvious or strongly implied.

- Bestow a fifth point if the CoT provides a complete, correct, and concise line of reasoning that leads efficiently to the final answer. It should include all necessary intermediate steps and be of appropriate length—neither too brief to omit key logic nor too long to include redundant or tangential content—reflecting expert-level clarity, precision, and efficiency.

User: \{question\}

<response>\{response\}</response>

After examining the user's instruction and the response:

- Briefly justify your total score, up to 100 words. Focus on relevance, logical completeness, clarity, and whether only the necessary reasoning steps were included without redundancy.

- Conclude with the score using the format: “Score: <total points>”

Remember to assess from the AI Assistant perspective, utilizing web search knowledge as necessary. To evaluate the response in alignment with this additive scoring model, systematically attribute points based on the outlined criteria.

\end{tcolorbox}
\caption{LLM-as-a-judge prompt format.}
\label{fig:llm-judge-prompt}
\end{figure*}

\section{Examples of Streamlined CoT}\label{sec:examples-binary}
We show qualitative examples of how our method removes redundant reasoning steps while preserving essential logic and the final answer. \colorbox{green!10}{The green part is the CoT after binary cutting,} \colorbox{blue!10}{the blue part marks the restored steps in backtracking,} \colorbox{red!10}{while the red part is removed redundant part.}

\begin{figure*}[htbp]
\centering
\small
\begin{tcolorbox}[colback=gray!00, 
                  colframe=black, 
                  width=16cm, 
                  arc=1.5mm, auto outer arc,
                  left=0.9mm, right=0.9mm,
                  boxrule=0.9pt, colbacktitle = black!65!black
                 ]
                 
\textbf{Question} \\
A \textit{normal magic square} of order $n$ is an arrangement of the integers from $1$ to $n^2$ in a square such that the $n$ numbers in each row, each column, and each of the two diagonals sum to a constant, called the \textit{magic sum} of the magic square. Compute the magic sum of a normal magic square of order $8$.

\textbf{Response of DeepSeek-R1} \\
\colorbox{green!10}{\parbox{\dimexpr\linewidth-2\fboxsep}{%
Okay, so I need to find the magic sum of a normal magic square of order 8. Hmm, let me think. A magic square is a square grid where the numbers in each row, column, and both main diagonals add up to the same number, called the magic sum.
}}
\colorbox{blue!10}{\parbox{\dimexpr\linewidth-2\fboxsep}{%
 Since it's a normal magic square, it uses the numbers from 1 to \( n^2 \), where \( n \) is the order. In this case, \( n \) is 8, so the numbers go from 1 to 64.
}}
\colorbox{red!10}{\parbox{\dimexpr\linewidth-2\fboxsep}{%
First, I remember there's a formula for the magic sum of a normal magic square. Let me try to recall. I think it involves the total sum of all the numbers in the square divided by the number of rows (or columns), since each row needs to add up to the magic sum. Let me check if that makes sense.

The total sum of numbers from 1 to \( n^2 \) can be calculated using the formula for the sum of the first \( m \) integers: \( \frac{m(m + 1)}{2} \). Here, \( m \) is \( n^2 \), so the total sum would be \( \frac{n^2(n^2 + 1)}{2} \). Then, since there are \( n \) rows, each row must add up to the total sum divided by \( n \). So the magic sum \( S \) should be \( \frac{n^2(n^2 + 1)}{2n} \). Simplifying that, we get \( S = \frac{n(n^2 + 1)}{2} \).

For example, a 3x3 magic square. The magic sum should be 15. Plugging \( n = 3 \) into the formula: \( \frac{3(3^2 + 1)}{2} = \frac{3(9 + 1)}{2} = \frac{3 \times 10}{2} = 15 \). That checks out. What about a 4x4? The magic sum should be 34. Let's see: \( \frac{4(4^2 + 1)}{2} = \frac{4(16 + 1)}{2} = \frac{4 \times 17}{2} = 34 \).

Therefore, for \( n = 8 \), substituting into the formula: \( S = \frac{8(8^2 + 1)}{2} \). First, calculate 8 squared: \( 8 \times 8 = 64 \). Then add 1: \( 64 + 1 = 65 \). Multiply by 8: \( 8 \times 65 = 520 \). Then divide by 2: \( \frac{520}{2} = 260 \). 

Wait, just to be thorough, let me recast the problem another way. Instead of using the formula, maybe I can compute the sum of numbers from 1 to 64 and then divide by 8. Let's see. The sum from 1 to 64 is \( \frac{64 \times 65}{2} = \frac{4160}{2} = 2080 \). Then divide by 8 (since there are 8 rows) to get the magic sum: \( \frac{2080}{8} = 260 \). Same result.

A sanity check: in a magic square, the magic sum is also equal to the number of the square (\( n \)) multiplied by the average of the first and last numbers. Since the numbers run from 1 to \( n^2 \), the first number is 1 and the last is \( n^2 \). Their average is \( \frac{1 + n^2}{2} \). So magic sum should be \( S = n \times \frac{1 + n^2}{2} \). Wait, that's the same as the formula we had earlier: \( S = \frac{n(n^2 + 1)}{2} \). Yep, that's exactly the same. So plugging in \( n = 8 \): \( S = 8 \times \frac{1 + 64}{2} = 8 \times \frac{65}{2} = \frac{8}{2} \times 65 = 4 \times 65 = 260 \). Perfect, same answer.

So all these different approaches lead to the same result, 260. Therefore, I can be confident that the magic sum for an 8x8 normal magic square is 260.

**Final Answer**  
The magic sum of a normal magic square of order 8 is \( \boxed{260} \).}}

\end{tcolorbox}
\caption{It can be observed that with proper CoT prefix, the SLM can infer $260$ without explicit calculation}
\label{fig:binary-cutting-example-2}
\end{figure*}

\begin{figure*}[htbp]
\centering
\small
\begin{tcolorbox}[colback=gray!00, 
                  colframe=black, 
                  width=16cm, 
                  arc=1.5mm, auto outer arc,
                  left=0.9mm, right=0.9mm,
                  boxrule=0.9pt, colbacktitle = black!65!black
                 ]
                 
\textbf{Question} \\
Today is January 31, 2015, welcome to the 2015 "Spring Cup" Final. The calculation result of the expression \( \frac{2015/1 + 2015/0.31}{1 + 0.31} \) is:

\textbf{Response of DeepSeek-R1} \\
\colorbox{green!10}{\parbox{\dimexpr\linewidth-2\fboxsep}{%
Okay, so I need to calculate this expression: \( \frac{2015/1 + 2015/0.31}{1 + 0.31} \). Hmm, let me break this down step by step. 

First, let's look at the numerator: \( \frac{2015}{1} + \frac{2015}{0.31} \). The denominator is \( 1 + 0.31 \). Maybe I can simplify each part separately before combining them. 

Starting with the numerator. The first term is \( 2015 \div 1 \), which is straightforward. Dividing any number by 1 gives the number itself, so that's just 2015. The second term is \( 2015 \div 0.31 \). Hmm, dividing by a decimal. Let me think how to compute that. 
}}
\colorbox{blue!10}{\parbox{\dimexpr\linewidth-2\fboxsep}{%
Dividing by 0.31 is the same as multiplying by \( \frac{1}{0.31} \). Let me calculate \( 1 \div 0.31 \). Well, \( 0.31 \times 3 = 0.93 \), and \( 0.31 \times 3.2 \) would be 0.992. So \( 1 \div 0.31 \) is approximately 3.2258. But I should probably do this more accurately. Let's do the division: \( 1 \div 0.31 \). 

Alternatively, I can write 0.31 as a fraction, which is \( \frac{31}{100} \). So \( 1 \div \frac{31}{100} = \frac{100}{31} \). So \( \frac{2015}{0.31} = 2015 \times \frac{100}{31} \). Let me compute that. 

\( 2015 \times 100 = 201500 \). Then, \( \frac{201500}{31} \). Let me do that division. \( 31 \times 6000 = 186000 \). Then \( 201500 - 186000 = 15500 \). \( 31 \times 500 = 15500 \). So total is \( 6000 + 500 = 6500 \). So \( \frac{2015}{0.31} = 6500 \)? Wait, really? Let me check that again. 
}}
\colorbox{red!10}{\parbox{\dimexpr\linewidth-2\fboxsep}{%
\( 31 \times 6500: 31 \times 6000 = 186000 \), \( 31 \times 500 = 15500 \). Adding them together gives \( 186000 + 15500 = 201500 \). Wait, but \( \frac{2015}{0.31} = 2015 \times \frac{100}{31} = \frac{201500}{31} \). Wait, \( 31 \times 6500 = 201500 \)? Wait, \( 31 \times 6500 \). Let me compute \( 31 \times 6000 = 186000 \) and \( 31 \times 500 = 15500 \). Then \( 186000 + 15500 = 201500 \). So yes, \( 31 \times 6500 = 201500 \). But \( 2015 \times 100 = 201500 \). So \( \frac{201500}{31} \) is indeed 6500. So that part is correct. Therefore, \( \frac{2015}{0.31} = 6500 \). 

So the numerator is \( 2015 + 6500 \). Let's add those together. \( 2000 + 6500 = 8500 \), and then the remaining 15 makes it 8515. So the numerator is 8515. 

Now the denominator is \( 1 + 0.31 = 1.31 \). So the entire expression simplifies to \( \frac{8515}{1.31} \). 

Hmm, so \( 8515 \div 1.31 \). Let's see if we can compute that. Maybe there's a simplification here. Let me check if \( 1.31 \times 6500 = 8515 \). Let me compute \( 1.31 \times 6500 \). 

First, \( 1 \times 6500 = 6500 \). Then \( 0.31 \times 6500 \). Let me compute \( 0.3 \times 6500 = 1950 \). Then \( 0.01 \times 6500 = 65 \). So \( 1950 + 65 = 2015 \). Then \( 6500 + 2015 = 8515 \). So yes, \( 1.31 \times 6500 = 8515 \). 

Therefore, \( \frac{8515}{1.31} = 6500 \). 

Wait a minute, so the entire expression simplifies to 6500? That seems too straightforward. Let me recap to check for any mistakes. 
Original expression: \( \frac{2015/1 + 2015/0.31}{1 + 0.31} \) 
Numerator: \( 2015 + \frac{2015}{0.31} = 2015 + 6500 = 8515 \) 
Denominator: 1.31 
\( \frac{8515}{1.31} = 6500 \) 

Yes, that seems correct. The key was recognizing that both the numerator and the denominator could be related through the same factor. Alternatively, maybe factor out 2015 from the numerator? Let me try that approach to verify. 

Numerator: \( 2015 \left( \frac{1}{1} + \frac{1}{0.31} \right) = 2015 \left( 1 + \frac{100}{31} \right) = 2015 \left( \frac{31 + 100}{31} \right) = 2015 \left( \frac{131}{31} \right) \) 

Denominator: \( 1.31 = \frac{131}{100} \) 

So the entire expression is \( \frac{2015 \cdot \frac{131}{31}}{\frac{131}{100}} = 2015 \cdot \frac{131}{31} \cdot \frac{100}{131} = 2015 \cdot \frac{100}{31} = \frac{201500}{31} = 6500 \). 

Same result. So that confirms it. Therefore, the answer is 6500. 
But just to make sure, let's do the division \( 8515 \div 1.31 \) step by step. 
First, write \( \frac{8515}{1.31} \). To eliminate the decimal, multiply numerator and denominator by 100 to get \( \frac{851500}{131} \). 

Now compute \( \frac{851500}{131} \). Let's see how many times 131 goes into 851500. 

But earlier, we saw that \( 131 \times 6500 = 851500 \). Because \( 131 \times 100 = 13100 \), so \( 131 \times 6500 = 131 \times (65 \times 100) = (131 \times 65) \times 100 \). Wait, \( 131 \times 65 \). Let's compute that. 

\( 131 \times 60 = 7860 \) and \( 131 \times 5 = 655 \). So \( 7860 + 655 = 8515 \). So \( 131 \times 65 = 8515 \). Therefore, \( 131 \times 6500 = 8515 \times 100 = 851500 \). 
Therefore, \( \frac{851500}{131} = 6500 \). 
So everything checks out. Therefore, the result of the expression is 6500. 
Final Answer: \( \boxed{6500} \)
}}
\end{tcolorbox}
\caption{It can be observed that the model has completed the inference before the original step "Let me check that again.", and the subsequent parts become redundant for the actual answer.}
\label{fig:binary-cutting-example-1}
\end{figure*}

\end{document}